\documentclass[11pt,a4paper]{article}
\usepackage[hyperref]{acl2020}
\usepackage{times}
\usepackage{latexsym}
\usepackage{microtype}
\usepackage{hyperref}
\usepackage{amsmath}
\usepackage{color}
\usepackage{graphicx}
\usepackage{makecell}
\usepackage{amsfonts}
\usepackage{multicol}
\usepackage{xspace}
\usepackage{multirow}

\usepackage{microtype}

\aclfinalcopy 

\title{Towards Robustness and Diversity: Continual Learning in Dialog Generation with Text-Mixup and Batch Nuclear-Norm Maximization}

\author{
Zihan Wang$^{1*}$, Jiayu Xiao$^{2}$ \thanks{~~These authors contributed equally to this work}, Mengxiang Li$^{1}$, Zhongjiang He$^{1}$, Yongxiang Li$^{1}$,\\ \textbf{Chao Wang}$^{1\dagger}$, \textbf{Shuangyong Song}$^{1}$\thanks{~~Corresponding Authors.}~~~~~~~ \\
$^{1}$China Telecom Corporation Ltd., Beijing, China \\
$^{2}$University of Chinese Academy of Sciences, Beijing, China \\
\texttt{\{wangzh54,hezj,liyx25,wangc17,songshy\}@chinatelecom.cn} \\
\texttt{jiayu.xiao@vipl.ict.ac.cn} \\
\texttt{limengx@126.com}
}

\date{}

\begin{document}
\maketitle

\begin{abstract}
In our dynamic world where data arrives in a continuous stream, continual learning enables us to incrementally add new tasks/domains without the need to retrain from scratch. A major challenge in continual learning of language model is catastrophic forgetting, the tendency of models to forget knowledge from previously trained tasks/domains when training on new ones. This paper studies dialog generation under the continual learning setting. We propose a novel method that 1) uses \textit{Text-Mixup} as data augmentation to avoid model overfitting on replay memory and 2) leverages Batch-Nuclear Norm Maximization (BNNM) to alleviate the problem of mode collapse. Experiments on a $37$-domain task-oriented dialog dataset and DailyDialog (a $10$-domain chitchat dataset) demonstrate that our proposed approach outperforms the state-of-the-art in continual learning.

\end{abstract}
 
\label{intro}

\section{Introduction}
In real dialog applications, it is crucial for systems to learn patterns for new tasks/domains without forgetting those relevant to previously trained tasks/domains. Even with strong generative prior~\cite{xiao2023r,lv2024pick}, deep neural networks still suffer from catastrophic forgetting when optimized on a sequence of tasks/domains ~\citep{cf}. Prior work has explored regularization\cite{ewc}, parameter isolation\cite{packnet,pathNet} and replay memory\cite{gem,agem,lamol} to address this issue. Among them, replay-based approaches have proven to be the most effective empirically \cite{replaybest}. Replay-based approaches work by storing a small set of representative exemplars from previous tasks/domains and including them with the current task/domain during training. However, this method has three major drawbacks. Firstly, replay memory merely provides partial signal for the prior task/domain, especially when the memory size is small. Deep neural networks tend to overfit a small replay memory by simply memorizing the replay exemplars, making it difficult to generalize well on previous task/domain examples outside the replay memory. Second, catastrophic forgetting occurs during the model adaption~\cite{xiao2022few,liu2023text} to the most recent task/domain, resulting in mode collapse\cite{mode_collapse} (i.e. different inputs from the previous tasks/domains and the latest task/domain are mapped into homogeneous dense feature representations). Finally, the samples in a batch are highly imbalanced -- most of the samples are from the current task/domain, while only a few are from the replay memory. Thus, the representation of samples from the replay memory  become similar to those of the current task/domain, aggravating the problem of mode collapse. 

To address the challenge of overfitting, we utilize \textit{Mixup}\cite{mixup} to augment our replay memory via linear interpolation between replay exemplars and current task/domain data. \textit{Mixup} regularizes the model by forcing it to behave linearly between training examples leading to better generalization. Although \textit{Mixup} has been used to alleviate catastrophic forgetting\cite{mixup_cl,mixup_cl2,mixup_cl3}, it has primarily been restricted to discriminative models\cite{wang2021towards} in computer vision. To the best of our knowledge, our application of \textit{Mixup} for language generation tasks under the continual learning setting is novel. In addition, to alleviate the issue of mode collapse, we leverage Batch Nuclear-Norm Maximization (BNNM)\cite{bnnm} to learn a higher-ranked feature space, which results in more diverse feature representation, and thus alleviate mode collapse issue. BNNM is also a remedy for the imbalanced data issue -- by improving the representation diversity of a batch, BNNM makes the representation of replay exemplars more discriminative from the current task data. We show that these two interventions, both separately and together, reduce the severity of forgetting in dialog generation when trained on new tasks/domains.

Overall, our contributions are three fold: ($i$) We propose \textit{Text-Mixup} for language generation, linearly interpolating between examples from previous tasks/domains and the current task/domain to avoid overfitting replay memory, and mitigate catastrophic forgetting.
($ii$) We encourage the model to learn a higher-ranked feature space with Batch Nuclear-Norm Maximization (BNNM) to improve representation diversity among tasks/domains, reducing mode collapse. ($iii$) Experimental results on a $37$-domain task-oriented dialog dataset \cite{madotto-2021-continual} and DailyDialog \cite{dailydialog}, a $10$-domain chitchat dataset, demonstrate the strength of our approach, outperforming state-of-the-art continual learning methods for language generation.

\section{Related Work}

\textbf{Continual learning} approaches can be roughly grouped into three categories. First, \textit{replay-based} methods maintain a collection of exemplars from previous tasks/domains in replay memory. The most straightforward way is to simply combine replay exemplars with the current task/domain when optimizing the model. Gradient Episodic Memory (GEM) extends this by explicitly constraining gradient updates so that the loss of replay exemplars never increases\cite{gem}. A-GEM simplifies GEM by reducing the number of task-specific constraints to one\cite{agem}. LAMOL is a generative-replay method designed specifically for language which simultaneously learns to both solve tasks and generate pseudo training samples\cite{lamol}. Despite these improvements, all of these methods suffer from both overfitting the replay memory and mode collapse\cite{mode_collapse}. 
The second category contains \textit{regularization-based} methods which impose constraints on network weights in order to preserve its performance on old tasks/domains. For example, Learning without Forgetting (LwF) \cite{LwF} retains knowledge of previous tasks/domains by knowledge distillation; Elastic Weight Consolidation (EWC) penalizes changes to important parameters (as estimated by Fisher information) during training \cite{ewc}; Memory Aware Synapses (MAS) \cite{MAS} redefines the parameter importance measure to an unsupervised setting. Methods of the final category, \textit{parameter isolation}, dedicate different model parameters to each task \cite{pathNet,packnet,HAT}, leading to significant increases in model size which preclude a fair comparison with our approach. Therefore, we do not consider parameter isolation in this paper.

\textbf{Data augmentation} is a widely-used technique to regularize deep networks for better generalization by forcing the model to learn task/domain appropriate invariance properties \cite{dataaug1,dataaug2}. Discrete data augmentation methods in NLP include word deletion, reordering, replacement, and addition \cite{dataaug_replace,dataaug_nlp}. However, recent work has shown that applying augmentation to intermediate dense representations outperforms discrete input augmentation in NLP. For example, SIMCSE\cite{simcse} demonstrates that dropout acts as a minimal form of augmentation, producing state-of-the-art results on semantic textual similarity using contrastive learning. Another class of data augmentation techniques, inspired by \textit{Mixup} \cite{mixup}, interpolates the input and labels of several real samples. \cite{textmix} was the first work to apply \textit{Mixup} to NLP by mixing hidden representations. Nevertheless, to the best of our knowledge, \textit{Mixup} has not been used to address catastrophic forgetting in language generation.

\textbf{Dialog Generation}. \textit{Task-Oriented Dialog} systems help people with many tasks like booking flights, checking the weather, making restaurant reservations, etc. They typically contain four modules: Natural Language Understanding, Dialog State Tracking, Dialog Policy, and Natural Language Generation (NLG). \textit{Chitchat} is another kind of dialog task for modeling conversation about daily life involving both the exchange of information and social bonding\cite{dailydialog}. For the task of generating natural responsed conditioned on dialogue acts, GPT-2 was used by \cite{scgpt} to improve few-shot learning, and by \cite{arper} in continual learning setting. The focus of this paper is on the NLG module that produces natural language utterances expressing a desired dialog act or response, a necessary component in both \textit{Task-Oriented Dialog} and \textit{Chitchat}.

\section{Methods}

\subsection{Problem Formalization}

For \textit{Task-Oriented Dialog}, our NLG component takes dialog acts as input and generates natural language responses. Formally, dialog act $X$ is the combination of intent $I$ and slot-value pairs $\{(s_i,v_i)\}_{i=1}^K$. The intents are system actions generated by the dialog-policy module such as \textit{Inform}, \textit{Request}, \textit{Confirm}, etc. Slot-value pairs record the category $s_i$ and corresponding content $v_i$ that need to be expressed in the response. The NLG module translates $X$ into natural language utterance $Y$. As for \textit{Chitchat}, the input $X$ is the dialog context of $t$ recent turns between two speakers. The NLG module generates $Y$, the next utterance.  

\subsection{Language Model}

In this paper, GPT-2 is used as semantically-conditioned neural language model. Given the dialog context $X = \{x_1, ..., x_m\}$ and output $Y = \{y_{1}, ..., y_{n}\}$, the conditional probability can be computed as:
\begin{equation}
p_{\theta}(Y|X) = \prod_{i=1}^{n} p_{\theta}(y_i|y_{<i}, X)
\end{equation}
Model parameter $\theta$ is optimized by maximizing the log-likelihood over the entire training set $D$ with $N$ samples. The loss $L_{\theta}$ is defined as:
\begin{equation}
L_{\theta}(D) = -\sum_{j=1}^N \sum_{i=1}^{n} logp_{\theta}(y_i^{(j)}|y_{<i}^{(j)},X^{(j)})
\end{equation}
At inference time, the model autoregressively generate output response given input. 

\subsection{Text-Mixup Augmentation on Replay Memory}
\label{sec:mixup}
\textit{Memory replay} has shown great success in mitigating catastrophic forgetting. Replay memory stores exemplars for previous tasks/domains which are then combined with data from the current task/domain to update model parameters. However, language models tend to overfit by simply memorizing the replay exemplars. Thus, it is helpful to regularize by performing data augmentation on the replay memory. 

\textit{Mixup} \cite{mixup} is a data augmentation technique that generates virtual training samples via linear interpolation between real training samples. It regularizes the model by encouraging it to behave linearly among the training data, enabling the model to learn smooth decision boundaries.

In contrast with vanilla \textit{Mixup}, we generate Mixup samples using dense intermediate representations. Specifically, for Transformer decoder-only language models with $L$ layers, we choose the hidden layer $l$ to perform \textit{Mixup}, where $l$ is an integer uniformly sampled from $[0, L]$ (note that the $0$-th layer is the embedding layer). Then we sample two examples, $\mathbf{x_i}$ from the current task and $\mathbf{x_j}$ from the replay memory. $h_l^i, h_l^j$ are their respective dense hidden representations from the $l$-th layer of the language model. We interpolate between these hidden representations to produce $\hat{h_l}$ which is passed to the subsequent layers of languauge model:

\begin{equation}
\hat{h_l} = \lambda h_l^i + (1-\lambda) h_l^j
\end{equation}

where $\lambda$ is sampled from a Beta distribution and $\alpha$ controls the degree of interpolation:

$$\lambda \sim Beta(\alpha,\alpha), \alpha \in (0,\infty)$$
$$\lambda = max(\lambda, 1-\lambda)$$

The labels at time step $t$ are mixed using the same ratio $\lambda$:
$$\hat{\mathbf{x}^t} = \lambda \mathbf{x}_i^t + (1-\lambda) \mathbf{x}_j^t, ~~~~t \in [1,n+m]$$

where $\mathbf{x}_i^t$ and $\mathbf{x}_j^t$ are one-hot vectors indicating the expected tokens at that timestep. 

The merits of \textit{Text-Mixup} are two-fold: first, as a data augmentation method, \textit{Text-Mixup} generates virtual samples based on samples from both the current task/domain and the stored exemplars, thus expanding the previously limited  replay memory to mitigate catastrophic forgetting. Moreover, the interpolated examples also approximate a regularized loss minimization \cite{mixup_how}. It imposes regularization on our model by forcing linear interpolations of hidden representations to have linear interpolations of output targets. Section \ref{sec:cl_comp} shows that \textit{Text-Mixup} consistently outperforms other text augmentation methods. 

\subsection{Nuclear-norm Maximization of Representation Matrix}

The phenomenon of "catastrophic forgetting" can be partially attributed to the tendency of language models to overfit new tasks/domains (and thereby forget old tasks/domains). As the model overfits a new task/domain, it tends to map different inputs $\mathbf{x}^{(1)}$ and $\mathbf{x}^{(2)}$ from the previous task/domain into homogeneous dense feature representations $\mathbf{\tilde{z}}$. This phenomenon is called \textit{mode collapse}. When mode collapse happens in text generation, the model tends to produce text patterns that frequently appears in the final task/domain regardless of the input. Some examples are shown in blue in table 1 (the final domain is "\textit{sgd\_train}"). We notice that the model often produces patterns that frequently appears in the final domain regardless of the input.  
 
Let us consider a transformer-based text generator that maps discrete text $\mathbf{x}$ into dense representations $\mathbf{z} \in \mathbb{R}^H$, where $H$ is the dimension of the dense feature space. For a mini-batch with $B$ instances including current task/domain data and replay exemplars, the text generator maps it to feature representation matrix $\mathbf{Z} \in \mathbb{R}^{B \times H}$. Mode collapse occurs when different rows of $\textbf{Z}$ are similar to each other. This is especially detrimental if the model maps replay exemplars into the same region as the current task/domain data, resulting in catastrophic forgetting. As matrix rank measures the number of linearly independent vectors in a matrix, $rank(\mathbf{Z})$ could be an indicator of mode collapse severity. Therefore, maximizing $rank(\mathbf{Z})$ should improve the overall representation diversity of a batch and thereby alleviate the problem of mode collapse. 
 
However, the maximization of matrix rank is known to be NP-hard so we cannot efficiently maximize the rank of matrix $\mathbf{Z}$ directly. As a workaround, we use nuclear norm \footnote{A matrix's nuclear norm is the sum of its singular values} maximization as a surrogate for maximizing matrix rank. \cite{rank} proves that a matrix's nuclear norm is the tightest convex envelop of its rank function within the unit ball. Mathematically,

\textbf{Theorem 1. (Fazel02)} \textit{For} $f(\mathbf{X}) = rank(\mathbf{X})$ \textit{over the set}
$S := \{\mathbf{X} \in \mathbb{R}^{m\times n} ~|~ ||\mathbf{X}||_F \leq 1\}$,
\textit{the convex envelope of} $f$ \textit{is the nuclear norm} $||\mathbf{X}||_{\star}$. \footnote{$||\mathbf{X}||_F$ stands for the Frobenius-norm of matrix $\mathbf{X}$}

\textbf{Proposition 1. (Fazel02)} \textit{For $f(\mathbf{X}) = rank(\mathbf{X})$ over the set} 
$S := \{\mathbf{X} \in \mathbb{R}^{m\times n} ~|~ ||\mathbf{X}||_F \leq M\}$,
\textit{the convex envelope of }$f$ is $\frac{1}{M}||\mathbf{X}||_{\star}$,\textit{ for} $M > 0$.

We can fix the Frobenius-norm of our batch representation matrix $\mathbf{Z}$ by performing L2 normalization on its rows. The Frobenius-norm of this normalized matrix is a constant with a known value which serves as our upper bound $M$:
\begin{equation}
||\mathbf{Z}||_F = \sqrt{\sum_{i=1}^B \sum_{j= 1}^H |Z_{i,j}|^2} = \sqrt{B}
\end{equation}

Thus, according to proposition $1$, the convex envelope of $rank(\mathbf{Z})$ is $||\mathbf{Z}||_{\star} / \sqrt{B}$ which is proportional to $||\mathbf{Z}||_{\star}$. By maximizing $||\mathbf{Z}||_{\star}$, we can encourage representational diversity within each batch and thereby mitigate mode collapse.  

We investigate applying BNNM at both the sentence-level and the token-level. At the token-level, we use the contextual representations of all tokens in the last layer to construct matrix $\mathbf{Z} \in R ^{BL \times H}$, where $L$ is the maximum sequence length. At the sentence-level, we employ a self-attentional pooling mechanism to generate sentence representations. This pooling is implemented as in \cite{clip}, where the query is the global average-pooled last layer hidden state and the keys and values are last layer hidden states. This results in a sentence representation matrix $\mathbf{Z} \in \mathbb{R}^{B \times H}, $where $H$ is the hidden size of the decoder-only language model. In both cases, the loss function we optimize for batch nuclear-norm maximization is:
\begin{equation}
L_{BNNM} = -\frac{1}{B} ||\mathbf{Z}||_{\star}
\end{equation}

\cite{grad} provide a derivation of the gradient of the nuclear-norm, so $L_{BNNM}$ is differentiable. We optimize the sum of $L_\theta$ and $L_{BNNM}$ weighted with hyperparameter $\kappa$:

\begin{equation}
L_{tot} = L_\theta + \kappa L_{BNNM}
\end{equation}\

The time complexity of computing the batch nuclear-norm is identical to the singular value decomposition. For a matrix $\mathbf{Z} \in \mathbb{R}^{B \times H} $, it is $O(min(B^2H, BH^2))$. As the batch size $B$ is always small, the cost is negligible.

The work most similar to ours is \cite{bnnm} which uses batch nuclear-norm maximization to improve both prediction discriminability and diversity. However, there are two major differences: (1) their work only performs experiments on classification tasks while we investigate the language generation task in continual learning setting. To the best of our knowledge, we are the first to use BNNM in language generation, and also the first one in the continual learning. (2) They apply BNNM to the output logits matrix while we demonstrate that applying BNNM to intermediate dense representations also boosts model performance.

\begin{table*}[h]
    \begin{center}
        \resizebox{1.0\hsize}{!}{
           \begin{tabular}{|c|c|}
            \hline
           \multicolumn{ 2}{|c|}{\textbf{Domain: sgd\_alarm}} \\
           \hline
           Dialog Act &  \textit{CONFIRM ( new\_alarm\_time = "3:45 pm" ; new\_alarm\_name = "Leave for home")}\\
           Reference & Okay, please confirm to make a new alarm at 3:45 pm in the name Leave for home. \\
           EWC & Okay, I have a \textcolor{blue}{train} that will leave for home at 3:45 pm. \textcolor{blue}{Would you like me to book it?}\\
           LAMOL & please confirm: you want to \textcolor{blue}{book} a new alarm for leave for home at 3:45 pm. \\
          
           Text-Mixup & please confirm the details: Creating an alarm at 3:45 pm with name Leave for home \textcolor{green}{and leave at 3:45 pm.} \\
           TM\_BNNM & Please confirm the details: make a new alarm for 3:45 pm called Leave for home. \\
           
           \hline
            \multicolumn{ 2}{|c|}{\textbf{Domain: sgd\_calendar}} \\
            \hline
            Dialog Act & \textit{OFFER ( event\_date = "next Thursday" ; available\_start\_time = "8 pm" ; available\_end\_time = "midnight")}\\
            Reference & Sure, next Thursday you have an empty slot from 8 pm until midnight. \\
            EWC & I have a 8 pm \textcolor{blue}{train that leaves} at 8 pm on next Thursday. \textcolor{blue}{Would you like me to book that?} \textcolor{red}{\textbf{\textit{\% miss {avaliable\_end\_time = "midnight"}}}}\\
            LAMOL & I have 1 available time for that day on next Thursday at 8 pm. \textcolor{blue}{do you need a ticket?} \textcolor{red}{\textbf{\textit{\% miss {avaliable\_end\_time = "midnight"}}}}\\
           
            Text-Mixup & There are \textcolor{orange}{7 empty slots on your calendar}. You have a empty slot from 8 pm until midnight on next Thursday. \\
            TM\_BNNM & You are all set up for next Thursday at 8 pm until midnight\\
        \hline
        \multicolumn{ 2}{|c|}{\textbf{Domain: TM2019\_restaurant}} \\
           \hline
        Dialog Act & \textit{restaurant\_reservation(time\_reservation\_accept="7:30";name\_restaurant\_accept="woodhouse fish company";num\_guests\_accept="4")} \\
        Reference & Ok, I have made a reservation for 7:30 tonight at woodhouse fish company, table for 4 with 1 booster seat, is that correct? \\
        EWC &  you will have a 7:30 \textcolor{blue}{flight} \textcolor{green}{with} woodhouse fish company for 4 people." \\
        LAMOL & Okay, so you will be \textcolor{blue}{leaving} at 7:30 am \textcolor{green}{with} woodhouse fish company, you have 4 people in your party. \textcolor{blue}{Have a safe trip.}" \\
        Text-Mixup & Alright, you are \textcolor{green}{looking for 4 people} at woodhouse fish company at 7:30 a.m? " \\
        TM\_BNNM & Okay. I have booked a table for you at 7:30 at woodhouse fish company for 4 people."\\

    \hline
    \end{tabular}
         }
    \caption{Examples of generated utterance by different methods. Errors are shown in colors. \textcolor{blue}{blue} texts indicates catastrophic forgetting, i.e. the model produces patterns that frequently appears in the latest domain regardless of the input. \textcolor{red}{red} denotes information omission. \textcolor{green}{green} means non-fluent generation, and \textcolor{orange}{orange} indicates hallucination.
    }
    \end{center}
\label{tab:example}
\end{table*}

\begin{table*}
\centering
\begin{tabular}{lcccc|cccc}
\hline
\multirow{2}{*}{Method} & \multicolumn{4}{c}{TOD37}                & \multicolumn{4}{c}{DailyDialog}               \\ \cline{2-9} 
                        & BLEU$\uparrow$ & TER$\downarrow$ & BS.$\uparrow$ & ME.$\uparrow$ &BLEU$\uparrow$ & TER$\downarrow$ & BS.$\uparrow$ & ME.$\uparrow$ \\ \hline

None & $21.62$& $0.882$& $0.889$& $0.221$&  $2.258$ & $1.198$ & $\mathbf{0.852}$ & $0.063$\\
Synonym substitution & $21.36$ &$0.957$ & $0.889$& $0.232$&$2.230$ & $1.198$ &$0.851$ & $0.064$\\
Word deletion & $21.85$ & $0.950$ & $0.891$ & $0.231$ & $2.170$ & $1.191$ & $\mathbf{0.852}$ & $0.062$ \\
Word insertion& $22.25$ & $0.942$ & $0.891$ & $0.237$ & $2.247$ &$1.222$ &  $\mathbf{0.852}$ & $0.064$ \\
Word swap  &  $20.84$ & $0.974$ & $0.888$ & $0.230$ & $2.284$ & $1.201$ & $0.851$ & $0.063$ \\
Dropout & $22.48$ & $0.942$ & $\mathbf{0.892}$ & $0.240$ & $2.193$ & $1.228$ & $0.851$ &  $0.064$ \\
Text-mixup & $\mathbf{23.29^*}$ & $\mathbf{0.857^*}$ & $\mathbf{0.892}$ & $\mathbf{0.246^*}$ & $\mathbf{2.342^*}$ & $\mathbf{1.197}$ & $\mathbf{0.852}$ & $\mathbf{0.065}$ \\
\hline
\end{tabular}
\caption{Comparison between \textit{Text-Mixup} and baseline data augmentation methods on a replay memory that contains 5 training samples for each domain. BS. stands for BertScore, and ME. stands for METEOR. For \textit{Text-Mixup}, the controllable factor $\alpha$ is set to be 0.6. Best results are \textbf{bold}. Mark * denotes the significantly better results with the significance level $p < 0.05$, comparing with other data augmentation methods.}
\label{table:aug}
\end{table*}

\begin{table*}
\centering
\begin{tabular}{lcccc|cccc}
\hline
\multirow{2}{*}{Method} & \multicolumn{4}{c}{TOD37}                & \multicolumn{4}{c}{DailyDialog}               \\ \cline{2-9} 
                        & BLEU$\uparrow$ & TER$\downarrow$ & BS.$\uparrow$ & ME.$\uparrow$ &BLEU$\uparrow$ & TER$\downarrow$ & BS.$\uparrow$ & ME.$\uparrow$ \\ \hline

Finetune & $20.16$ & $0.886$ & $0.883$ & $0.195$ & $1.414$ & $1.215$ & $0.845$ & $0.044$ \\
Replay & $21.62$ & $0.882$ & $0.889$ & $0.221$ &  $2.258$ & $1.198$ & $\mathbf{0.852}$ & $0.063$ \\
EWC & $21.97$ & $0.850$ & $0.883$ & $0.195$ & $1.977$ & $1.291$ & $0.849$ & $0.061$ \\
Lamol & $19.93$ & $0.915$ & $0.883$ & $0.201$  & $2.126$ & $1.207$ &$\mathbf{0.852}$ & $0.062$ \\
AGEM & - & - & - & - & $2.187$ & $1.178$ & $0.851$ & $0.061$ \\
Text-mixup & $23.29$ & $0.857$ & $\mathbf{0.892}$ & $0.246$ & $2.342$ & $1.197$ & $\mathbf{0.852}$ & $\mathbf{0.065}$ \\
TM\_BNNM (token-level) & $23.34$ & $0.856$ & $\mathbf{0.892}$ & $0.245$ & $2.145$ & $1.246$ & $0.851$ & $0.062$ \\
TM\_BNNM (sent.-level) & $\mathbf{23.59^*}$ & $\mathbf{0.849^*}$ & $\mathbf{0.892}$ & $\mathbf{0.247}$ & $\mathbf{2.515^*}$ & $\mathbf{1.167^*}$ & $\mathbf{0.852}$ & $0.061$ \\
\hline
MULTI & $24.13$ & $0.838$ & $0.898$ & $0.258$ & $7.228$ & $1.012$ &   $0.867$ & $0.095$ \\
\hline
\end{tabular}
\caption{Comparison between TM\_BNNM and other state-of-the-art continual learning methods. AGEM is not implemented for TOD37 due to memory limitation. Best results except for MULTI are \textbf{bold}. Mark * denotes the significantly better results with the significance level $p < 0.05$, comparing with other methods.}
\label{table:cl}
\end{table*}

\section{Experiments}

We evaluate our approach on a 37-domain task-oriented dialog dataset, \textbf{TOD37} \cite{madotto-2021-continual}, and 10-domain chitchat dataset, \textbf{DailyDialog} \cite{dailydialog}. Both of the datasets are written in English. The statistics of both the datasets are provided in Appendix ~\ref{sec:appendix}. We adopt the methodology described in \cite{cl}: after a task/domain is trained, the training data of that task/domain is no longer accessible except in replay memory. After all tasks/domains are trained, the final model is scored using each task/domain's test set. 

\subsection{Experimental Setup}

Our pretrained backbone is GPT-2 which contains $12$ layers with $12$ attention heads per layer and has a hidden size of $768$, with parameter size of 117M. The input to our model is tokenized with pretrained BPE codes. For DailyDialog, we select the last $t = 3$ turns as context in order to generate the next utterance. For TOD37 and DailyDialog, the sequences are truncated or padded to the length of $80$ and $100$ respectively. Other hyperparameters are listed in Table \ref{tab:hyper}. We optimize our model with AdamW \cite{adamw} with $\beta_1 = 0.9, \beta_2 = 0.999$, and $\epsilon = 1e-08$. 

\begin{table}
\centering
\begin{tabular}{r|rr}
\hline & \textbf{TOD37} & \textbf{DailyDialog} \\ \hline
batch size & $32$ & $128$ \\
epochs & $1$ & $5$ \\
learning rate & $6.25e-05$ & $6.25e-04$ \\
$\alpha$ & $0.6$ & $0.6$ \\
$\kappa$ & $0.4$ & $0.6$ \\
\hline
\end{tabular}

\caption{Hyperparameter configuration for TOD37 and DailyDialog.}
\label{tab:hyper}
\end{table}

During decoding, for TOD37, we generate $k = 5$ utterances via nucleus sampling with $p = 0.9$ and choose the utterance with the lowest slot error rate (ERR) as our final output. ERR is the ratio of missing slots in a generated utterance to ground truth slots in the input dialog act \cite{err}. For DailyDialog, we generate a single utterance via nucleus sampling with $p = 0.9$. 

All experiments were conducted on five random permutations of task/domain orders. The results reported are averaged across these permutations. The experiments were implemented and evaluated on $1$ NVIDIA GeForce 3090 GPU using PyTorch.

\subsection{Metrics}

We use a variety of automatic metrics to evaluate the quality of the generated text. Specifically, we calculate BLEU \cite{bleu} \footnote{We implemented BLEU evaluation the same as \url{https://github.com/moses-smt/mosesdecoder/blob/master/scripts/generic/multi-bleu.perl}}, METEOR \cite{meteor} \footnote{We use METEOR version 1.5: \url{http://www.cs.cmu.edu/~alavie/METEOR/README.html}}), and TER \cite{ter} \footnote{We use TER version 0.7.25: \url{http://www.cs.umd.edu/~snover/tercom/}}) to measure word overlap and BertScore \cite{bertscore} \footnote{We use BertScore version 0.3.12} to measure semantic overlap between our generated text and the reference text.  

\section{Results}

\subsection{Data Augmentation}
\label{sec:cl_comp}

We compare \textit{Text-Mixup} against several common  augmentation techniques for text. For discrete data  baselines, we perform random perturbation including word deletion, insertion, substitution and swap on $10\%$ of the words in each sentence. We use the rule-based discrete data augmentation method described in \cite{eda}\footnote{\url{https://github.com/jasonwei20/eda_nlp}}. As for continuous data-augmentation baselines, we apply dropout to the embedding, self-attention and fully-connected layers by simply passing the same sentence to the pretrained GPT-2 model twice, as in SIMCSE \cite{simcse}. The dropout rate is set $0.1$. For \textit{Text-Mixup}, each mini-batch sampled from the replay memory is augmented via mixing with a mini-batch sampled from the current task/domain as described in Section \ref{sec:mixup}.

All the data augmentation methods are used to augment a small replay memory that contains $5$ training samples for each domain. As shown in Table ~\ref{table:aug}, the proposed text-mixup method consistently outperforms all discrete and continuous text data augmentation baselines. \textit{Text-Mixup} successfully retains useful knowledge from prior tasks/domains while preventing the model from overfitting on the limited replay memory, thus mitigating catastrophic forgetting. We note that \textit{Text-Mixup} is complementary to other data augmentation methods -- applying \textit{Text-Mixup} together with any of these other methods would likely lead to improved results. However, we leave that for future work.

\begin{table*}
\centering
\begin{tabular}{lcccc|cccc}
\hline
\multirow{2}{*}{Method} & \multicolumn{4}{c}{TOD37}                & \multicolumn{4}{c}{DailyDialog}               \\ \cline{2-9} 
                        & BLEU$\uparrow$ & TER$\downarrow$ & BS.$\uparrow$ & ME.$\uparrow$ &BLEU$\uparrow$ & TER$\downarrow$ & BS.$\uparrow$ & ME.$\uparrow$ \\ \hline

w/o data augmentation & $21.62$& $0.882$& $0.889$& $0.221$&  $2.258$ & $1.198$ & $\mathbf{0.852}$ & $0.063$\\
Text-mixup ($\alpha = 0.4$) & $23.00$ & $0.864$ & $0.890$ & $0.243$ & $2.191$ & $1.190$ & $0.851$ & $0.061$\\
Text-mixup ($\alpha = 0.6$)& $\mathbf{23.29}$ & $\mathbf{0.857}$ & $\mathbf{0.892}$ & $0.246$ & $\mathbf{2.342}$ & $\mathbf{1.197}$ & $\mathbf{0.852}$ & $\mathbf{0.065}$ \\
Text-mixup ($\alpha = 0.8$)& $22.98$& $0.863$& $0.891$& $\mathbf{0.247}$ & $2.280$& $1.208$& $0.851$&$0.063$\\
\hline
\end{tabular}
\caption{ Parameter sensitivity for $\alpha$ in text-mixup. BS. stands for BertScore, and ME. stands for METEOR. Best results are \textbf{bold}. }
\label{tab:mixup_ablation}
\end{table*}

\begin{table*}
\centering
\begin{tabular}{lcccc|cccc}
\hline
\multirow{2}{*}{Method} & \multicolumn{4}{c}{TOD37}                & \multicolumn{4}{c}{DailyDialog}               \\ \cline{2-9} 
                        & BLEU$\uparrow$ & TER$\downarrow$ & BS.$\uparrow$ & ME.$\uparrow$ &BLEU$\uparrow$ & TER$\downarrow$ & BS.$\uparrow$ & ME.$\uparrow$ \\ \hline
with Text-Mixup &  $23.59$ & $0.849$ & $0.892$ & $0.247$ & $2.515$ & $1.167$ & $0.852$ & $0.061$\\
w/o Text-Mixup & $22.54$ & $0.862$ & $0.890$ & $0.232$ &  $2.357$ & $1.172$ & $0.852$ & $0.062$ \\

\hline
\end{tabular}
\caption{Ablation study on TM\_BNNM with and without Text-Mixup. }
\label{table:BNNM_ablation}
\end{table*}

\begin{table*}
\centering
\begin{tabular}{lcccc|cccc}
\hline
\multirow{2}{*}{Method} & \multicolumn{4}{c}{TOD37}                & \multicolumn{4}{c}{DailyDialog}               \\ \cline{2-9} 
                        & BLEU$\uparrow$ & TER$\downarrow$ & BS.$\uparrow$ & ME.$\uparrow$ &BLEU$\uparrow$ & TER$\downarrow$ & BS.$\uparrow$ & ME.$\uparrow$ \\ \hline

$\kappa = 0.0$ & $23.29$ & $0.857$ & $\mathbf{ 0.892}$ & $0.246$ & $2.342$ & $1.197$ & $\mathbf{0.852}$ & $\mathbf{0.065}$ \\
$\kappa = 0.2$ & $23.32$ & $0.852$ & $0.891$ & $0.245$ & $2.300$ &  $1.238$ & $0.850$ & $0.063$ \\
$\kappa = 0.4$ & $\mathbf{23.59}$ & $\mathbf{0.849}$ & $\mathbf{0.892}$ & $\mathbf{0.247}$ & $2.323$ & $1.172$ & $0.851$ & $0.057$ \\
$\kappa = 0.6$ & $23.21$ & $0.854$ & $0.890$ & $0.245$ & $\mathbf{2.515}$ & $\mathbf{1.167}$ & $\mathbf{0.852}$ & $0.061$ \\
$\kappa = 0.8$ & $23.19$ & $0.853$ & $0.891$ & $0.243$ & $2.403$ & $1.180$ & $0.851$ & $0.060$ \\
\hline
\end{tabular}
\caption{ Parameter sensitivity for $\kappa$ in batch nuclear norm maximization. BS. stands for BertScore, and ME. stands for METEOR. Best results are \textbf{bold}. }
\label{table:kappa}
\end{table*}


\subsection{Continual Learning}

We compare \textit{Text-Mixup} with batch nuclear-norm maximization (TM\_BNNM) against several continual learning baselines:

\begin{itemize}
\item \textbf{EWC} \cite{ewc} regularizes parameters important for previous tasks/domains to mitigate forgetting. The importance of parameters is measured via the Fisher Information Matrix. We set the combining ratio $\lambda$ to $0.01$. 

\item \textbf{LAMOL} \cite{lamol} is a generative-rehearsal method that learns a model to generate pseudo rehearsal samples while learning to solve tasks. 

\item \textbf{A-GEM} \cite{agem} imposes constraints on gradient updates, ensuring that at every training step the loss over the replay memory does not increase. 

\item The \textbf{Replay} method optimizes the model with current task/domain data and exemplars of previous tasks/domains selected by \textit{Herding}. In this paper, for all methods that use replay memory, we retain $5$ exemplars for each task/domain.  

\item \textbf{Finetune}: in this setting, we train the model without any regularization or replay. 

\item \textbf{Multi}: in this setting, we train the model on all tasks/domains together. This is the performance "upper-bound" in continual learning. 

\end{itemize}

\begin{figure}[t]
  \centering
   \includegraphics[width=0.99\linewidth]{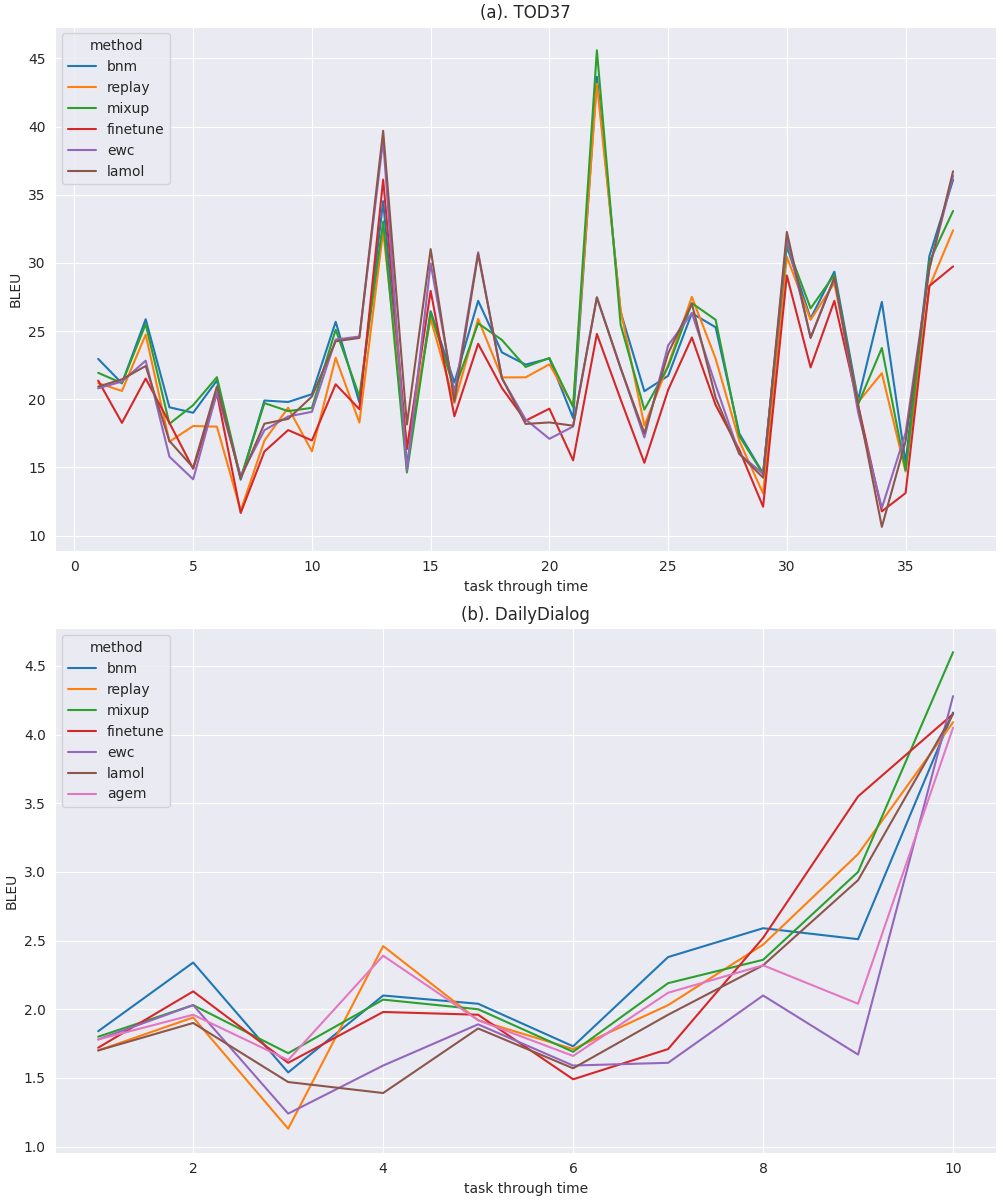}
   \vspace{-2mm}
   \caption{The performance of different continual learning methods: BLEU scores on each task/domain according to its position in the training curriculum after training on all tasks/domains sequentially.
   }
   \label{fig:cl}
   \vspace{-6mm}
\end{figure}

Table~\ref{table:cl} shows that TM\_BNNM with sentence-level batch nuclear-norm maximization outperforms other state-of-the-art continual learning methods. Figure ~\ref{fig:cl} plots the BLEU score for each continual learning method on TOD37 and DailyDialog. We note that our proposed method performs better than other continual learning methods on the first few tasks, demonstrating its effectiveness at alleviating catastrophic forgetting. 
\begin{table*}
\centering
\begin{tabular}{lc|c}
\hline
\multirow{2}{*}{Method} & TOD37 & DailyDialog               \\
                        & \% unique n-grams  & \% unique n-grams \\ \hline
with BNNM & $14.86$/$41.69$/$63.73$/$78.24$ & $30.00$/$70.13$/$86.87$/$92.30$ \\
w/o BNNM & $14.82$/$41.31$/$63.11$/$77.63$& $28.69$/$68.84$/$85.82$/$91.54$ \\

\hline
\end{tabular}
\caption{The percentage of unique uni/bi/tri/quad-grams generated by our method with and without BNNM. }
\label{table:ngram}
\end{table*}

\subsection{Ablation Study}

We conduct ablation experiments to better understand the properties of each of the two components that comprise our proposed method. 

Firstly, for \textit{Text-Mixup}, Table \ref{tab:mixup_ablation} shows model performance under different settings of the hyperparameter $\alpha$. We note that \textit{Text-Mixup} performs the best on both datasets when $\alpha = 0.6$. Table \ref{table:BNNM_ablation} shows that removing \textit{Text-Mixup} leads to poorer result, indicating that \textit{Text-Mixup} contributes to the model performance. 

\begin{figure}[t]
  \centering
   \includegraphics[width=0.99\linewidth]{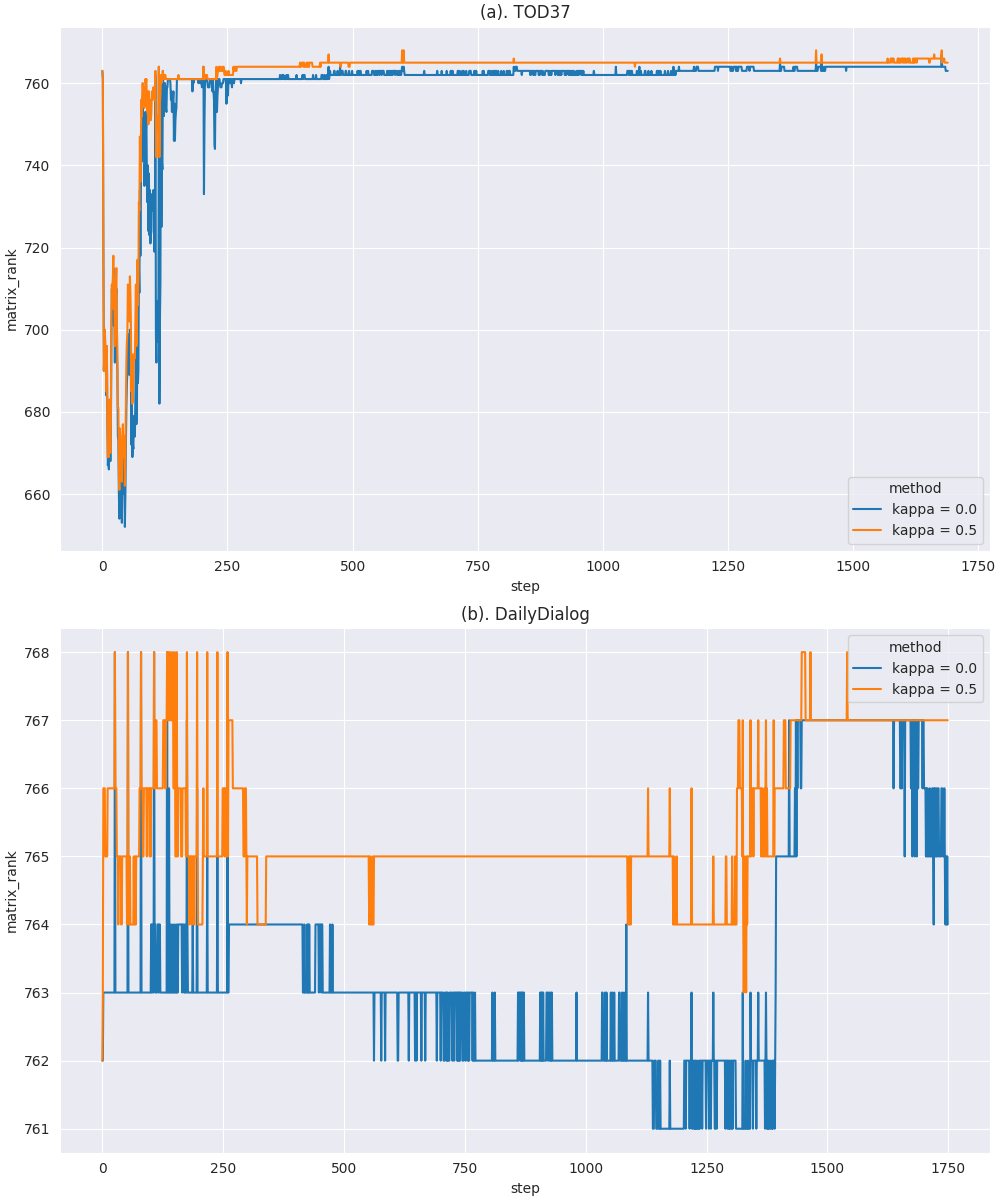}
   \vspace{-2mm}
   \caption{The rank of representation matrix with and without BNNM during the whole training process.
   }
   \label{fig:bnm}
   \vspace{-6mm}
\end{figure}

Secondly, Figure \ref{fig:bnm} shows the rank of example representation matrices with and without token-level BNNM during the training process. We can see that maximizing batch nuclear norm effectively maximizes matrix rank, thus improving representation diversity. Moreover, table \ref{table:ngram} shows that BNNM also improve n-gram diversity in our generated text, suggesting that the model more effectively avoid mode collapse and retains prior domains. Table \ref{table:kappa} shows the parameter sensitivity for our BNNM weighting hyperparameter $\kappa$. We note that when $\kappa = 0.4$, BNNM improves BLEU by $+0.2$ for TOD37, and when $\kappa = 0.6$, BNNM improves  BLEU score by $+0.173$ for DailyDialog. 

\section{Conclusions}
This work introduces TM\_BNNM, a continual learning method for addressing catastrophic forgetting in dialog generation task. TM\_BNNM utilizes a novel \textit{Text-Mixup} method as data augmentation to improve the generalization ability of model, and leverage Batch Nuclear-Norm Maximization to alleviate the mode collapse problem. The experiment results have shown that it outperforms state-of-the-art method in task-oriented dialog generation and chitchat generation datasets.

\bibliography{acl2020}
\bibliographystyle{acl_natbib}

\appendix
\section{Appendix}
\label{sec:appendix}

\begin{table*}
\centering

\begin{tabular}{l|cccccc}
\hline\hline

Domain &  Train & Val & Test & \makecell[c]{Avg. Turns\\ Per Dialog}&\makecell[c]{Avg. Words \\ Per Dialog}& \makecell[c]{Avg. Words \\ Per Utterance} \\
\hline

Ordinary Life & 3090 & 284 & 272 & 8.25 & 109.21 & 12.35\\
School Life & 410 & 34 & 43 & 9.38 & 130.46 & 13.00\\ 
Culture \& Education & 44 & 4 & 7 &10.14 & 134.4 & 12.34 \\
Attitude \& Emotion & 563 &50 &54 &6.42 & 92.71 & 13.59\\
Relationship & 3733 & 337 & 321 &7.34 &107.87 & 13.82 \\
Tourism& 890& 81& 94 & 7.94 & 104.49 & 12.28\\
Health & 224& 23&20 &9.70 & 140.32 & 13.56\\
Work & 1648& 145& 131&7.48 & 128.33 & 16.28\\
Politics& 109& 9&13 & 11.67 & 122.97 & 9.62\\
Finance& 407& 33&45 & 8.73 & 154.69 & 16.81 \\

\hline

\end{tabular}
\caption{Statistics of different domains in DailyDialog dataset.}
\label{tab:dataset_dialy}
\end{table*}

\begin{table*}
\centering

\begin{tabular}{l|cccccc}
\hline\hline

Domain &  Train & Val & Test &  \makecell[c]{Avg. Actions \\ Per Sample} &  \makecell[c]{Avg. Words \\ Per Sample} \\
\hline

TM19\_movie & 3010 & 367 & 345 & 1.0 & 24.91 \\
TM19\_auto & 2127 & 224 & 284 & 1.0 & 26.63 \\
TM19\_restaurant & 2585 & 331 & 334 & 1.0 & 27.02 \\
TM19\_pizza & 1329 & 172 & 174 & 1.0 & 27.67 \\
TM19\_uber & 2420 & 291 & 279 & 1.0 & 22.61 \\
TM19\_coffee & 1384 & 152 & 189 & 1.0 & 26.91 \\
TM20\_flight & 12064 & 1552 & 1463 & 1.0 & 29.11 \\
TM20\_food-ordering & 2395 & 278 & 288 & 1.0 & 25.68\\
TM20\_hotel & 7826 & 997 & 990 & 1.0 & 36.94\\
TM20\_music & 4223 & 540 & 528 & 1.0 & 21.53\\
TM20\_restaurant & 8633 & 1099 & 1030 & 1.0 & 28.55\\
TM20\_sport & 12059 & 1554 & 1545 & 1.0 & 23.99\\
TM20\_movie & 10869 & 1404 & 1247 & 1.0 & 33.65\\
MWOZ\_taxi & 396 & 69 & 59 & 1.0 & 26.01\\
MWOZ\_train & 579 & 62 & 62 & 1.0 & 29.5\\
MWOZ\_restaurant & 3373 & 116 & 158 & 1.0 & 34.53\\
MWOZ\_hotel & 1706 & 179 & 218 & 1.0 & 32.82\\
MWOZ\_attraction & 353 & 36 & 34 & 1.06 & 32.96\\
sgd\_restaurants & 2254 & 200 & 342 & 1.0 & 34.53\\
sgd\_media & 1174 & 194 & 400 & 1.13 & 21.41\\
sgd\_events & 3779 & 490 & 1071 & 1.0 & 34.65\\
sgd\_music & 1714 & 255 & 531 & 1.0 & 31.73\\
sgd\_movies & 1760 & 217 & 80 & 1.22 & 24.5\\
sgd\_flights & 4538 & 370 & 1187 & 1.0 & 46.34\\
sgd\_ridesharing & 366 & 46 & 111 & 1.0 & 26.85\\
sgd\_rentalcars & 1128 & 172 & 344 & 1.13 & 38.23\\
sgd\_buses & 1584 & 204 & 384 & 1.03 & 35.46\\
sgd\_hotels & 2361 & 312 & 866 & 1.0 & 33.32\\
sgd\_services & 2679 & 374 & 805 & 1.0 & 32.44\\
sgd\_homes & 1569 & 249 & 422 & 1.0 & 42.93\\
sgd\_banks & 592 & 71 & 143 & 1.0 & 22.6\\
sgd\_calendar & 507 & 67 & 148 & 1.15 & 31.5\\
sgd\_alarm & 219 & 28 & 51 & 1.11 & 26.77\\
sgd\_weather & 126 & 26 & 59 & 1.0 & 23.57\\
sgd\_travel & 199 & 23 & 42 & 1.0 & 29.62\\
sgd\_payment & 77 & 11 & 19 & 1.0 & 32.38\\
sgd\_trains & 205 & 51 & 84 & 1.0 & 32.52\\

\hline

\end{tabular}
\caption{Statistics of different domains in TOD37 dataset.}
\label{tab:dataset_tod}
\end{table*}

\end{document}